\documentclass[letterpaper]{article} 
\usepackage{aaai25}  
\usepackage{times}  
\usepackage{helvet}  
\usepackage{courier}  
\usepackage[hyphens]{url}  
\usepackage{graphicx} 
\usepackage{natbib}  
\usepackage{caption} 
\usepackage[linesnumbered,ruled,vlined]{algorithm2e}
\SetKwInput{KwInput}{Input}     
\SetKwInput{KwOutput}{Output}   
\usepackage{amsmath,amssymb,amsfonts}
\usepackage{amsthm}
\usepackage{accents}
\usepackage{graphicx}
\usepackage{subcaption}
\usepackage{booktabs}
\usepackage{multirow}
\usepackage{dsfont}
\usepackage{makecell}
\usepackage{threeparttable}

\newcommand{\ie}{{\em i.e.}}

\newcommand{\bfx}{{\bf{x}}}
\newcommand{\bfy}{{\bf{y}}}

\newtheorem{theorem}{Theorem}

\newtheorem{lemma}[theorem]{Lemma}

\title{A Black-Box Evaluation Framework for Semantic Robustness\\ in Bird's Eye View Detection}
\author{
    Fu Wang\textsuperscript{\rm 1, \rm 2}, 
    Yanghao Zhang\textsuperscript{\rm 2}, 
    Xiangyu Yin\textsuperscript{\rm 2}, 
    Guangliang Cheng\textsuperscript{\rm 2}, \\
    Zeyu Fu\textsuperscript{\rm 1}, 
    Xiaowei Huang\textsuperscript{\rm 2}, 
    Wenjie Ruan\textsuperscript{\rm 1,  \rm 2}
}
\affiliations{
    \textsuperscript{\rm 1} Department of Computer Science, University of Exeter, Exeter, EX4 4QF, UK\\
    \textsuperscript{\rm 2} Department of Computer Science, University of Liverpool, Liverpool, L69 3BX, UK\\
    \rm{\{fw377, z.fu, wruan\}@exeter.ac.uk}\\
    \rm{\{yanghao.zhang, x.yin22, guangliang.cheng, xiaowei.huang\}@liverpool.ac.uk}\\
}

\begin{document}

\maketitle

\begin{abstract}
Camera-based Bird's Eye View (BEV) perception models receive increasing attention for their crucial role in autonomous driving, a domain where concerns about the robustness and reliability of deep learning have been raised.
While only a few works have investigated the effects of randomly generated semantic perturbations, aka natural corruptions, on the multi-view BEV detection task, we develop a black-box robustness evaluation framework that adversarially optimises three common semantic perturbations: geometric transformation, colour shifting, and motion blur, to deceive BEV models, serving as the first approach in this emerging field.
To address the challenge posed by optimising the semantic perturbation, we design a smoothed, distance-based surrogate function to replace the mAP metric and introduce SimpleDIRECT, a deterministic optimisation algorithm that utilises observed slopes to guide the optimisation process.
By comparing with randomised perturbation and two optimisation baselines, we demonstrate the effectiveness of the proposed framework.
Additionally, we provide a benchmark on the semantic robustness of ten recent BEV models.
The results reveal that PolarFormer, which emphasises geometric information from multi-view images, exhibits the highest robustness, whereas BEVDet is fully compromised, with its precision reduced to zero.
\end{abstract}

\begin{links}
    \link{Code}{https://github.com/TrustAI/RobustBEV}
\end{links}

\section{Introduction}

The landscape of autonomous driving is undergoing a significant transformation, propelled by the rapid advancement of deep neural network-based methods~\cite{chitta2021neat,prakash2021transfuser,zhang2022beverse,hu2022mile,hu2022stp3,chen2022lav,HuYCLSZCDLWLJLD23}.
One of the central elements in this evolution is the Bird's Eye View (BEV) perception. 
As highlighted by recent studies~\cite{HuYCLSZCDLWLJLD23,xie2023robobev}, BEV perception is a critical intermediary in boosting the overall efficacy of the full autonomy stack. 
Significant research efforts have been directed towards developing BEV perception~\cite{ma2022vision}.
Particularly, camera-based BEV perception models have gained increasing attention over the LiDAR-based approaches in recent development. 
This shift is primarily attributed to the cost-effectiveness of camera systems compared to LiDAR systems.
Furthermore, camera-based perception models can provide rich semantic information that is not only meaningful for human observers but also enhances the performance in downstream tasks~\cite{HuYCLSZCDLWLJLD23}.

BEV detection is a fundamental perception task that identifies objects from multi-view input and maps them onto the BEV planform.
BEV detection models can extract useful features from the images obtained, which are then utilised by downstream tasks such as object tracking~\cite{yin2024dimba}, occupancy prediction, and route planning~\cite{HuYCLSZCDLWLJLD23}.
It is indisputable that autonomous driving is a safety-critical scenario~\cite{chowdhury2020attacks} that requires robust and reliable BEV detection models~\cite{CaoWXYFYCLL21}.
Recently, a number of studies~\cite{yang2022provable,zhu2023understanding,xie2023adversarial,xie2023robobev} have been conducted to evaluate the robustness of camera-based BEV detection models against adversarial attacks~\cite{carlini2019evaluating} and natural corruptions~\cite{hendrycks2019benchmarking}.
The findings from these studies, unfortunately, indicate that BEV models also suffer from adversarial threats.
In particular, many works investigated the robustness of the BEV detection models against random natural corruption~\cite{zhu2023understanding,xie2023robobev}.
These perturbations, crafted to mimic natural corruptions, account for potential disruptions stemming from environmental conditions and hardware constraints. 
Given that the generation of natural corruption is random and independent of the target models, these studies primarily shed light on the `average-case' robustness of BEV detection models against semantic perturbations~\cite{xie2023robobev}. 
However, these approaches may not fully expose the underlying vulnerabilities of BEV models if the `worst-case’ robustness is not taken into account.
Similar to the spirit of the adversarial examiner framework~\cite{shu2020identifying}, we move beyond the natural corruptions and study the adversarial robustness of BEV detection models when subjected to semantic perturbations and develop a gradient-free, query-based evaluation framework.
By leveraging advances in global optimisation techniques, the proposed evaluation framework is compatible with most existing camera-based BEV models. 
It can identify more competitive adversarial perturbations than natural corruption methods, thereby providing a more reliable measurement of the `worst-case' performance of BEV models.

This work begins by formulating the semantic perturbation threat model. 
We notice that adopting the mean Average Precision (mAP) metric as an objective function for optimisation presents challenges due to its discontinuous nature.
To address this issue, we design a distance-based surrogate objective function that incorporates the bounding box matching mechanism used in BEV detection tasks.
As defined in Eq.~\eqref{eqn:final_object_func}, our surrogate function is continuous for matched bounding boxes while exhibiting a strong negative correlation with detection precision.
These characteristics enable the function to serve as an effective proxy for optimising semantic perturbations aimed at degrading the performance of targeted BEV models.
Furthermore, we propose a query-based global optimisation algorithm called SimpleDIRECT, to solve the optimisation problem by finding the `worst-case' perturbation that compromises the target BEV models.
In this algorithm, we boost the optimisation performance and computational efficiency of the DIRECT optimisation~\cite{piyavskii1972algorithm} by reducing redundancies in identifying potential optimal solutions and leveraging slope information to guide the optimisation process.
To evaluate the semantic robustness of camera-based BEV detection models, we introduce an evaluation framework focusing on three common semantic perturbations: geometric transformation, colour shift, and motion blur. These perturbations represent typical challenges posed by camera-system anomalies and real-world driving scenarios. We conduct extensive experiments on the nuScenes dataset~\cite{caesar2020nuscenes} to assess the performance of recent BEV models under these challenging conditions and empirically demonstrate the superiority of our proposed evaluation framework over randomised perturbations used in previous literature and two optimisation-based baseline methods.
In our robustness benchmark on ten BEV models, our evaluation compromised the BEVDet~\cite{huang2021bevdet} that shows considerable resilience against natural corruption.
These findings highlight shortcomings in previous literature~\cite{zhu2023understanding,xie2023robobev} and the critical need for our work.

\section{Related Works}\label{sec:relate-work}
\paragraph{Robustness of BEV detection }\label{sec:related_work_robust}

In this work, we focus on the object detection task in BEV perception based on multi-view images.
As a fundamental task in autonomous driving, the robustness of object detection has attracted notable research interest~\cite{kong2023robo3d,kong2023robodepth,xie2023robobev}.
To assess the robustness of camera-based BEV detection models, several efforts have been made to study the pixel-level and patch-level adversarial attacks~\cite{AbdelfattahY0W21,ParkLCM21,ZhangLWWLJ22,xie2023adversarial}.
Unlike adversarial attacks~\cite{carlini2019evaluating}, which often exploit model vulnerabilities with unrealistic inputs, semantic perturbations~\cite{hendrycks2019benchmarking} mimic naturally occurring corruptions. 
Therefore, semantic perturbations have been widely considered in robustness evaluations to assess model performance under real-world conditions~\cite{MirzaBJOOSB21,YuTXLLWCHWL23}.
Most recently, ~\citet{kong2023robo3d} and~\citet{xie2023robobev} introduced natural corruption respectively on LiDAR and camera-based BEV models, and~\citet{zhu2023understanding} studied the influence of common visual corruptions.
While these approaches provide valuable insights, they may not comprehensively reveal the underlying vulnerabilities of BEV models. Their limitation lies in relying on randomised perturbations to assess the semantic robustness of these models.

\paragraph{Deterministic Global Optimisation}
Originating from the Lipschitz optimisation~\cite{piyavskii1972algorithm}, DIRECT is a gradient-free deterministic optimisation algorithm that works on Lipschitz continuous objective functions but does not rely on the Lipschitz constant~\cite{Jones1993Lipschi,Gablonsky01,zhang2023model}.
Recent studies have applied DIRECT to assess the robustness of deep neural network models against geometric transformations~\cite{WangXRH23} and adversarial patches~\cite{xu2023sora}, proving its efficiency in approximating the global optimum and highlighting its effectiveness.

\section{Semantic Adversarial Threat Model}
\paragraph{Problem Formulation}
In the domain of 3D object detection utilising multi-monocular views, an input frame $\mathbf{x} \in \mathbb{R}^{N \times H \times W \times 3}$ encapsulates $N$ images, each with dimensions of height $H$, width $W$, and three colour channels.
Given an end-to-end detection model $\mathcal{F}$, we have $\mathcal{F}(\mathbf{x}) = \hat{\mathbf{y}}$, where $\hat{\mathbf{y}} \in \mathbb{R}^{U \times k}$ is the model's prediction that comprises $U$ bounding boxes, each described by a $k$-dimensional vector.
Considering a $s$-dimensional semantic perturbation, we apply the perturbation with different setups on each image in the input frame, resulting in a $sN$-dimensional parameter space.
We denote by $\mathcal{S}_\theta$ a semantic perturbation with parameter $\theta \in \mathbb{R}^{sN}$.
Given the annotation $\bfy \in \mathbb{R}^{V \times k}$ that contains $V$ ground-truth bounding boxes and $\hat{\mathbf{y}}_\theta$ the predicted bounding boxes from the perturbed example $\mathcal{S}_\theta(\bfx)$, we aim to find the optimal $\theta$ such that 
\begin{equation}\label{eqn:init_object_func}\textstyle
    \underset{\theta \in \Theta}{\arg\max}\; \mathcal{L}\big(\mathcal{F}(\mathcal{S}_\theta(\bfx)), \bfy\big),
\end{equation} 
where $\mathcal{L}$ is the detection loss and $\Theta$ is a bounded adversary space.
Furthermore, we assess the target model's robustness by determining the optimal perturbation for each frame in a video clip, aiming to measure its {\em worst-case} performance in the face of semantic perturbations within the autonomous driving scenario.

\paragraph{Distance-based Objective Function}

Building upon the framework established by the nuScenes dataset~\cite{caesar2020nuscenes}, 
a majority of existing works~\cite{xie2023robobev,zhu2023understanding,li2022bevformer} employs the NDS to measure the detection performance in BEV perception. 
The NDS is a composite measurement that combines five different metrics, with mean Average Precision (mAP) playing the most significant role. 
In the nuScenes detection task, we observe that mAP calculation differs from the convention. 
Instead of using the intersection over union metric, it is based on the 2D centre distances between predicted and annotated bounding boxes on the ground plane.
This implementation introduces a box-matching mechanism based on a distance threshold $\tau$~\cite{caesar2020nuscenes}.
Let $\mathcal{D}: \mathbb{R}^{U\times k} \times \mathbb{R}^{k} \to \mathbb{R}^{U}$ be a function that returns the 2D centre distances between predicted bounding boxes and annotated boxes.
Given a ground truth detection box $\bfy_v$ and predicted detection boxes $\hat{\bfy}$, the matched prediction exists if $\min \mathcal{D}(\hat{\bfy}^v, \bfy_v) < \tau$ and is given by $\arg\min \mathcal{D}(\hat{\bfy}^v, \bfy_v)$.
Once a match is identified, the corresponding detection box is recorded and subsequently excluded from further matching processes.
The box-matching mechanism on an annotation $\mathbf{y}$ can be formulated as:
\begin{equation}\textstyle\label{eqn:box_match}
    \sum_{v=1}^V \mathds{1}\big( \min \mathcal{D}(\hat{\bfy}^v, \bfy_v) \leq \tau\big),
\end{equation}
where $\mathds{1}$ is the indicator function and $\hat{\bfy}^v$ is a subset of $\hat{\bfy}$  containing the predicted boxes with the same classification label as the ground truth box $\bfy_v$.

Apparently, Eq.~\eqref{eqn:box_match} is not an ideal objective function for optimisation due to its discontinuity, so we propose the following loss function as a surrogate of the mAP metric:
\begin{equation}\textstyle\label{eqn:final_object_func}
  \mathcal{L}\big(F(S_\theta(\bfx)), \bfy\big) =  \sum_{v=1}^V \min\big(\min \mathcal{D}(\hat{\bfy}^v_{\theta}, \bfy_v), \tau\big),
\end{equation} 
which ensures that the function value remains continuous for pairs of predicted and ground truth boxes that meet the matching criteria given in Eq.~\eqref{eqn:box_match}.
Maximising Eq.~\eqref{eqn:final_object_func} would force the semantic perturbation to increase the distance between the matched bounding boxes and simultaneously avoid the emergence of new matches.

\section{Simplified DIRECT Optimisation}
DIRECT~\cite{Jones1993Lipschi} is a Deterministic Optimisation (DO) algorithm that has been adopted to identify the optimal semantic perturbation setup in the adversary space and achieved considerable performance~\cite{WangXRH23}.
In this section, we delve into its core mechanisms and propose a simplified strategy to enhance its performance.

\paragraph{Algorithm Overview}\label{sec:method-overview}
To locate the global optimum $\theta^\star$ in the adversarial space $\Theta$, the DO starts by projecting $\Theta$ into a unit hypercube.
The whole adversarial space $\Theta$ is then treated as the root node and divided into a tree-like partition throughout the optimisation process~\cite{Munos11soo}.
At every depth of this partition tree, we can form a set of subspaces $\vartheta_h$, where $0 \leq h\leq H$ and $H$ is the maximum depth.
Let the diameter of a node be $\delta(\Theta) = ||\Theta||_\infty$, the subspaces at the same depth possess an identical diameter $\delta$, \ie, $\delta(\Theta_{a})$ = $\delta(\Theta_{b})$ holds for any $ \Theta_{a}$, $\Theta_{b} \in \vartheta_h$.
The deterministic optimisation operates by selectively dividing specific leaf nodes at different depths, thereby creating new nodes, and it evaluates the objective function at the central point, denoted as $\theta$ of each new node.
The PO node selection and partitioning are conducted repeatedly during the optimisation loop.

As a specific implementation of DO, DIRECT divides the PO nodes following a space trisection strategy.
Without loss of generality, consider a $d$-dimensional node $\Theta_{r}$ at depth $h$ that is poised for further exploration.
$\Theta_{r}$ is shaped as a hyperrectangle characterised by $m$ dimensions each with longer sides of length $3^{-h}$ and $d - m$ dimensions each with shorter sides of length $3^{-h-1}$, where $0<m\leq d$.
DIRECT performs query sampling within those $m$ dimensions that correspond to long edges. 
The locations of sampled points are given by $\theta_{r} \pm 3^{-h-1} \mathbf{e}_i$, for $i \in \{1,\ldots,m\}$, where $\mathbf{e}_i$ is a unit vector along $i$-th dimension.
After evaluating the objective function at all newly sampled points, DIRECT divides each long side into three equal parts, creating new subspaces.
These subspaces are centred at the sampled points.
Crucially, points yielding optimal results are allocated proportionally larger subspaces within $\Theta_{r}$ during partition.
This principle ensures that areas of the search space with more promising results are explored more thoroughly~\cite{Jones1993Lipschi}.
Besides, \citet{WangXRH23} proposed to track the slopes between the queried points during space trisection.
To achieve this, they enabled DIRECT to compute the slopes along each long edge as follows:
\begin{equation*}\textstyle
    \hat{K}_{r,i}^+ = \frac{|\mathcal{L}(\theta_r)-\mathcal{L}(\theta_r^{i+})|}{3^{-h-1}}\quad\text{and}\quad\hat{K}_{r,i}^- = \frac{|\mathcal{L}(\theta_r)-\mathcal{L}(\theta_r^{i-})|}{3^{-h-1}},
\end{equation*}
where  $\theta_r^{i+}$ and $\theta_r^{i-}$ denote $\theta_r \pm 3^{-h-1} \mathbf{e}_i$, respectively.
Then the largest slope at node $\Theta_{r}$ given by \begin{equation*}\textstyle
\hat{K}_r = \max\{\hat{K}_{r,1}^+,\hat{K}_{r,1}^-\ldots,\hat{K}_{r,m}^+,\hat{K}_{r,m}^-\}.
\end{equation*}
We illustrate the space transition in Fig.~\ref{fig:po_partition}, highlighting potential slope computation points with dashed red lines.
The recorded slopes can be viewed as an approximate to the lower bound of local Lipschitz constants that assist for estimating the potential improvement at each node, so we also adopt this approach in our practice, which will be elaborated in the following sections.
We present the pseudocode of this deterministic optimisation pipeline in Alg.~\ref{alg:workflow} and refer the readers to~\cite{Jones1993Lipschi,Munos11soo} for more details.

\begin{algorithm}[t]\small
\DontPrintSemicolon
  \KwInput{The objective function $\mathcal{L}$, the parameter space $\Theta$, the number of iterations $T$, the maximum depth $H$}
  \KwOutput{The optimal perturbation factor $\tilde{\theta}$}
  Initialise $\Theta$ as the root node and let $\mathcal{P}=\{\Theta\}$.\;
  $t \gets 0$, $q \gets 0$, $\tilde{\theta} \to 0$\;

  \While{$(t \leq T) \& (\mathcal{P} \neq \varnothing)$}
  {
    Initialise $\mathcal{X} = \{\}$\;
    \For{each potential optimal node $\Theta_p$ in $\mathcal{P}$}
    {
       \If{ $\delta_p > 3^{-H}$}{
          \For{dimension $i$ with long edge of $\Theta_p$}
          {
            $\mbox{Append}(\mathcal{X}, \theta_p \pm \frac{1}{3}\delta_p\mathbf{e}_i)$\;
          }  
        }
    }
    $\mathcal{Y} = \mathcal{L}(\mathcal{X})$\;
    \If{$\max\;\mathcal{Y} > \mathcal{L}(\tilde{\theta})$}
        {$\tilde{\theta} = {\arg\max}\;\mathcal{Y}$}

    \For{each potential optimal nodes $\Theta_p$ in $\mathcal{P}$}
    {
        Trisect node $\Theta_p$ based on query results in $\mathcal{Y}$\;
        Update $\Theta_p$'s size $\delta_p$ and local slope $\hat{K}_{p}$\;
    }
    $\mathcal{P} = \text{NodeSelection()}$\;
   	t = t+1	\;}
\caption{Deterministic Optimisation Pipeline}\label{alg:workflow}
 \vspace{-2mm}
\end{algorithm}

\subsection{Redundancy in the PO Node Selection}\label{sec:redundant}
\begin{lemma}[Potential Optimal nodes~\cite{Gablonsky01}]~\label{lem:find_po}
Let $\mathcal{L}_{\max}$ denote the current best query result, and $H$ represent the depth of the partition tree.
Given a node set $\vartheta = \bigcup_{h=1}^{H}\vartheta_h$ and a positive tolerance $\epsilon > 0$. 
For any node $\Theta_p$ at depth $h$, we can define three sets as follows: $\mathcal{I}^h_1 = \{\Theta_q \in \vartheta: \delta_q < \delta_p\}$, $\mathcal{I}^h_2 = \{\Theta_q \in \vartheta:\delta_q > \delta_p\}$ and $\mathcal{I}^h_3 = \{\Theta_q \in \vartheta:\delta_q = \delta_p\}$, where $\delta_p = \delta(\Theta_p)$.
Then the node $\Theta_p$ is said to be potentially optimal if 
\begin{equation}\label{eqn:po_cond0}\textstyle
    \mathcal{L}(\theta_p) \geq \mathcal{L}(\theta_q), \forall \Theta_q \in \mathcal{I}^h_3,
\end{equation}
where $\theta_a$ denotes the centre of node $\Theta_a$,
and the following inequations hold:
\begin{equation}\label{eqn:po_cond1}\textstyle
\min _{\Theta_q \in \mathcal{I}^h_{1}} \frac{\mathcal{L}\left(\theta_{p}\right)-\mathcal{L}\left(\theta_{q}\right)}{\delta_{p}-\delta_{q}} \geq \max _{\Theta_q \in \mathcal{I}^h_{2}} \frac{\mathcal{L}\left(\theta_{q}\right)-\mathcal{L}\left(\theta_{p}\right)}{\delta_{q}-\delta_{p}};
\end{equation}
\begin{equation}\label{eqn:po_cond2}\textstyle
    {\left|\mathcal{L}_{\max }\right|}\cdot\epsilon \leq \mathcal{L}\left(\theta_{p}\right) - \mathcal{L}_{\max }- \delta_{p} \max _{\Theta_q \in \mathcal{I}^h_{2}} \frac{\mathcal{L}\left(\theta_{q}\right)-\mathcal{L}\left(\theta_{p}\right)}{\delta_{q}-\delta_{p}}.
 \end{equation}
\end{lemma}

\begin{figure}[!tb]
    \centering
    \begin{subfigure}[t]{0.2\textwidth}
        \includegraphics[width=\textwidth]{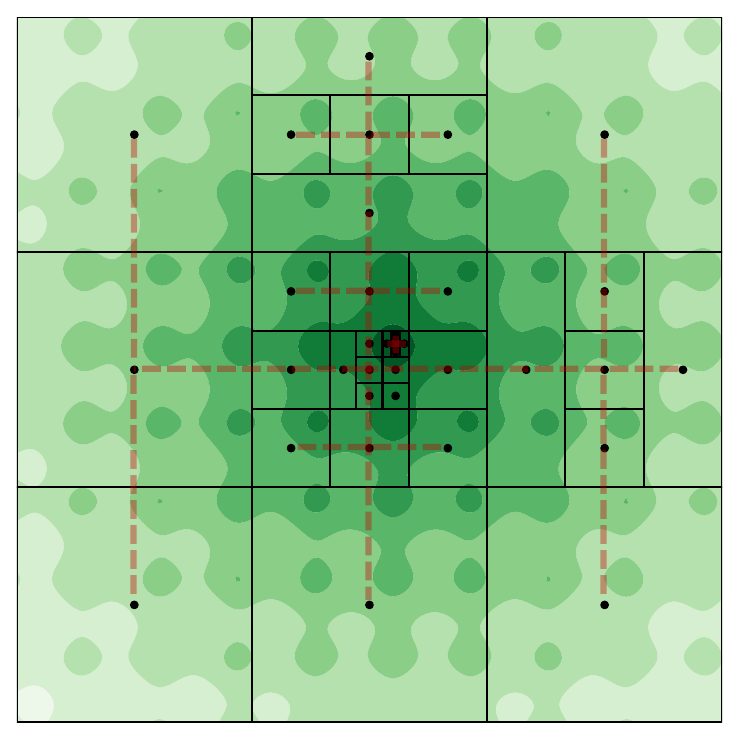}
        \caption{Space partition}
        \label{fig:po_partition}
    \end{subfigure} 
    \begin{subfigure}[t]{0.26\textwidth}
    \includegraphics[width=\textwidth]{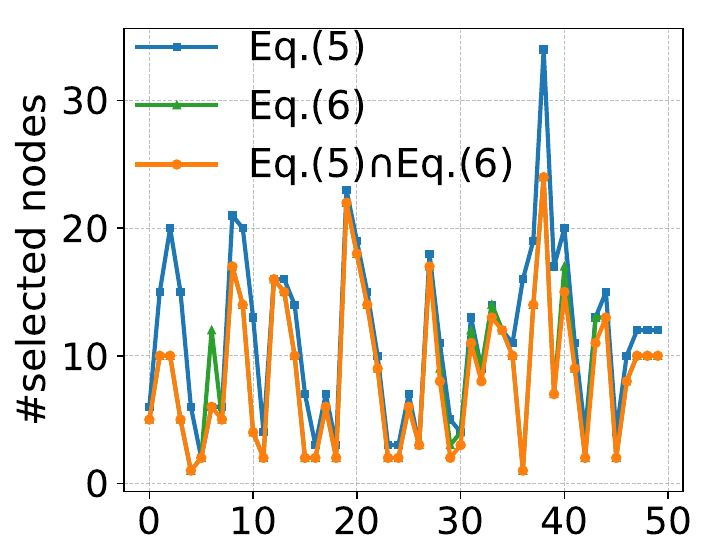}
    \caption{Counting the PO nodes}
    \label{fig:po_cond}
    \end{subfigure} 
    \begin{subfigure}[t]{0.3\textwidth}
        \includegraphics[width=\textwidth]{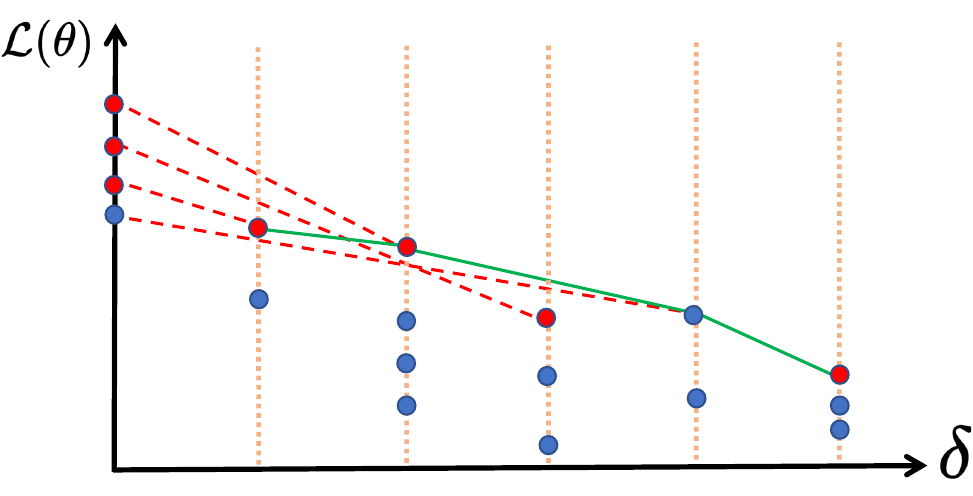}
        \caption{The difference between Eq.~\eqref{eqn:po_cond1} and ~\eqref{eqn:simple_cond}}
        \label{fig:simple_cond}
    \end{subfigure}
    \caption{Fig. (a) visualises the space trisection strategy, where the dashed red lines represent where the slopes can be obtained. Fig. (b) demonstrates the redundancy introduced by Eq.~\eqref{eqn:po_cond1}, which often qualifies notably more nodes than Eq.~\eqref{eqn:po_cond2}. Fig. (c) illustrates the difference between Eq.~\eqref{eqn:po_cond1} (green lines) and Eq.~\eqref{eqn:simple_cond} (dashed red lines).  }
    \label{fig:support_demo}
     \vspace{-2mm}
\end{figure}

The PO node selection is a core mechanism of DIRECT.
According to Lemma~\ref{lem:find_po}, DIRECT employs three conditions to select the PO nodes. 
While Eq.~\eqref{eqn:po_cond0} identifies nodes yielding the highest query result at each depth, Eq.~\eqref{eqn:po_cond1} and \eqref{eqn:po_cond2} refine the selection based on the potential for improvement over the current optimal results~\cite{Gablonsky01}
Although both Eq.~\eqref{eqn:po_cond1} and \eqref{eqn:po_cond2} play a critical role in selecting PO nodes, our analysis reveals that Eq.~\eqref{eqn:po_cond1} has less impact in identifying PO nodes compared to Eq.~\eqref{eqn:po_cond2}.
To illustrate, we applied DIRECT optimisation to the  6-dimensional Schwefel test function~\cite{yang2010engineering} and tracked the number of nodes satisfying each condition.
As shown in Fig.~\ref{fig:po_cond}, we can see that the PO nodes selection is mostly dominated by Eq.~\eqref{eqn:po_cond2}, while Eq.~\eqref{eqn:po_cond1} qualifies a notably larger number of nodes at each iteration, implying its relative inefficiency.

\begin{algorithm}[!t]\small
\DontPrintSemicolon
  \KwInput{ The maximum number of PO nodes $R$}
  \KwOutput{ A set of PO nodes $\mathcal{P}$}
  Set $\tilde{\vartheta} = \{\}$\;
  \For{each depth $h \in H$}
    {
      $\mbox{Append}(\tilde{\vartheta}, \Theta_p)$ \textbf{if} $\Theta_p$ satisfies Eq.~\eqref{eqn:po_cond0} and~\eqref{eqn:po_cond2}\;
    }
  \If{$|\tilde{\vartheta}| > R$}
  {
      Rank $\tilde{\vartheta}$ based on Eq.~\eqref{eqn:simple_cond}\;
      $\mathcal{P} = \{\mbox{Top }R-1\mbox{ nodes from } \tilde{\vartheta}\} \cup \{\arg\max{\delta(\tilde{\vartheta})}\}$
  }
  \Else
      {$\mathcal{P} = \tilde{\vartheta}$}
\caption{NodeSelection() in SimpleDIRECT}\label{alg:simpleDIRECT}
 \vspace{-2mm}
\end{algorithm}

\subsection{Simplifying the PO Node Selection}\label{sec:simple_po}

Querying BEV perception models is computationally expensive, and introducing less qualified and redundant PO nodes would slow down the optimisation process.
Due to the impractical selection condition given in Eq.~\eqref{eqn:po_cond1}, we propose to select the $R$ highest-ranking nodes in terms of potential improvement.
Specifically, consider $\Tilde{\vartheta}$ as a set of lead nodes qualified by Eq.~\eqref{eqn:po_cond0} and~\eqref{eqn:po_cond2}, if its cardinality $|\Tilde{\vartheta}| > R$, we assign a score $I$ for each selected node.
For any $\Theta_j \in \Tilde{\vartheta}$, the score is calculated as 
\begin{equation}\label{eqn:simple_cond}\textstyle
    I(\Theta_j) = \mathcal{L}(\theta_j) + 0.5\cdot \delta(\Theta_j)\cdot\hat{K}_j,
\end{equation}
where $I(\Theta_j)$ estimates the potential improvement at $\Theta_j$, taking into account its diameter $\delta(\Theta_j)$ and the largest slope $\hat{K}_j$ observed in its closest vicinity.
We then further refine $\Tilde{\vartheta}$ by picking the top $R-1$ leaf nodes based on this score given in Eq.~\eqref{eqn:simple_cond} and the leaf node with the largest diameter, determining the final PO nodes.
The leaf node with the largest diameter is selected because it always satisfies Eq.~\eqref{eqn:po_cond1} and Eq.~\eqref{eqn:po_cond2} and plays an important role in the convergence guarantee~\cite{Gablonsky01}.
The pseudocode of this simplified PO node selection strategy is presented in Alg.~\ref{alg:simpleDIRECT},
and, by integrating Alg.~\ref{alg:workflow} and Alg.~\ref{alg:simpleDIRECT} together, we propose an improvement of DIRECT, namely \textbf{SimpleDIRECT}.
An illustration of the difference between our strategy and its counterpart given in Eq.~\eqref{eqn:po_cond1} is shown in Fig.~\ref{fig:simple_cond}, where the nodes selected by Eq.~\eqref{eqn:po_cond1} is connected by green lines and our strategy is visualised by red dashed lines.
We highlight that Eq.~\eqref{eqn:po_cond1} selects the PO nodes based on an analytic comparison between nodes at different depths, while our strategy gives nodes with larger slopes higher priority and could reveal optimal nodes earlier.
Moreover, our simplified condition does not require the left-hand side of Eq.~\eqref{eqn:po_cond1}, whose computational complexity is $\mathcal{O}(H^2)$.
As the complexity of estimating and ranking the potential improvement is $\mathcal{O}(H \log H)$, our approach also advances Eq.~\eqref{eqn:po_cond1} in terms of computational efficiency.

\section{Applying Semantic Perturbations}
When deploying a trained BEV perception model in a vehicle equipped with cameras different from those used during the training data collection, variations in camera systems may result in \textbf{geometric transformations} in the captured monocular images.
Similarly, due to sensor variations among different camera systems, \textbf{colour shift} in recorded images is a common issue~\cite{xie2023robobev}.
On the other hand, as a natural and inevitable distortion in autonomous driving contexts, \textbf{motion blur} could significantly affect image clarity and, consequently, the accuracy of BEV perception models.
Therefore, we apply these three semantic perturbations to examine the robustness of BEV perception derived from multi-view images.

\paragraph{Geometric Transformation}
\label{sec:geo_perturb}

We evaluate the models' robustness against scaling and translation transformation.
For each monocular image, $\bfx_n \in \mathbb{R}^{H\times W\times C}$, and its corresponding variant, $\bfx_i^\prime$, 
the relationship between pixel coordinates in the perturbed image and the original image is governed by the intrinsic matrix. 
Specifically, a pixel in $\bfx_i^\prime$ at location $(\mathtt{x}_{i}^\prime,\mathtt{y}_{i}^\prime)$ corresponds to a pixel in $\bfx_n$ with index $(\mathtt{x}_{j},\mathtt{y}_{j})$.
Such a mapping can be written as
\begin{equation}\label{eqn:geo_perturb}\textstyle
\left[\begin{array}{l}
\mathtt{x}_{j} \\
\mathtt{y}_{j}
\end{array}\right]
=
\left[\begin{array}{lll}
\theta_s^{hor}& 0 \;\;& \theta_t^{hor} \\
0 \;\;& \theta_s^{vrt} \;\;& \theta_t^{vrt}
\end{array}\right] \left[\begin{array}{c}
\mathtt{x}_{i}^\prime \\
\mathtt{y}_{i}^\prime \\
1
\end{array}\right],
\end{equation}
where $\theta_s^{hor}$, $\theta_s^{vrt}$, $\theta_t^{hor}$, and $\theta_t^{vrt}$ are four intrinsic perimeters.
By manipulating $\theta_s^{hor}$ and $\theta_s^{vrt}$, we can respectively simulate horizontal and vertical scaling effects, while altering $\theta_t^{hor}$ and $\theta_t^{vrt}$ introduces horizontal and vertical translation in the image.
Furthermore, the geometric transformation can be formulated as a Lipschitz continuous operation, which has been proved by~\citet{LiWXRK00021} and~\citet{WangXRH23}.

\paragraph{Colour Shift}
\label{sec:colour_perturb}

Echoing the previous practices~\cite{mohapatra2020towards,zhang2022proa} in perturbing the colour space, we manipulate image colours in the HSB (Hue, Saturation, and Brightness) space rather than RGB.
Given an image, we first obtain its representation in the HSB colour space and then directly change the hue, saturation, and brightness values to achieve colour shifting.
Let $\bfx^{hue}$, $\bfx^{sat}$, and $\bfx^{brt}$ represent the colour channels in the HSB space. The hue, saturation, and brightness perturbations can formulated as follows:
\begin{align}\textstyle
    &S^{hue}(\bfx^{hue}, \theta^{hue}) = (\bfx^{hue} + \theta^{hue}) \bmod 2\pi,\\
    &S^{sat}(\bfx^{sat}, \theta^{sat}) =\min(\max(0, \theta^{sat} \cdot \bfx^{sat}), 1),\\
    &S^{brt}{(\bfx^{brt}, \theta^{brt})} = \min(\max(\bfx^{brt} + \theta^{brt}, 0), 1).
\end{align}
The perturbed image is then converted back to the RGB space to evaluate the robustness of the BEV perception models.
A rigorous proof of continuity for colour projection is beyond the scope of this paper. We assume that colour shifting is Lipschitz continuous for small perturbations, given the inherent continuity of colour mapping from HSB to RGB space~\cite{levkowitz1993glhs}.

\paragraph{Motion Blur}
\label{sec:motion_perturb}

To integrate motion blur into our robustness evaluation, we keep the size of the blur kernel a constant and optimise the blur angle $\theta^{ang}$ and direction $\theta^{dir}$ to maximise its impact on BEV perception models~\cite{eriba2019kornia}.
In this case, motion blur can be divided into two procedures: rotating the blur kernel at a given direction and applying the kernel in a convolution manner.
The continuity of rotating a 2-D metric has been proved by \citet{WangXRH23}, while \citet{LiangH21} demonstrated the convolution operation without bias is a linear transformation.
Therefore, the overall process of performing motion blur is continuous.

\section{Experiments}

In this section, we first demonstrate the effectiveness of the designed objective function and make initial observations on the impact of different semantic perturbations on the BEV perception models' performance.
Then, we evaluate the effectiveness of the proposed method in locating the optimal setup for different perturbations and compare its performance to baselines. Furthermore, we benchmark the robustness of recent BEV perception models, providing a comprehensive assessment of the robustness of the models under varied adversarial image corruptions.
Due to limited GPU resources, we adopt the validation set of the mini-subset of nuScenes~\cite{caesar2020nuscenes} and expand our approach to the whole validation set.
The implementation details are deferred to \textbf{Appendix}.

\setlength{\tabcolsep}{1mm}
\begin{table}[!t]\small
  \centering
    \begin{tabular}{cc}
    \toprule
    \multicolumn{1}{c}{Perturbation} & \multicolumn{1}{c}{Parameters for each image}  \\
    \midrule
    \multirow{2}[1]{*}{\makecell{Geometric\\Transformation} } &   $\theta_s^{vrt} \in [-\gamma H, \gamma H]$,  $\;\theta_s^{hor} \in [-\gamma W, \gamma W]$ \\
          &   $\theta_t^{vrt}, \theta_t^{hor}\in [1-\gamma, 1+\gamma ]$ \\
    \midrule
    \multirow{2}[0]{*}{Colour Shift} & $\theta^{brt} \in [-\gamma, \gamma]$, $\;\theta^{sat} \in [1-\gamma, 1+\gamma]$  \\
          & $\theta^{hue}\in[-\pi\cdot\gamma, \pi\cdot\gamma]$    \\
      \midrule
    Motion Blur &    $\theta^{ang}\in [-\pi,\pi], \theta^{dir} \in [-1, 1]$  \\
    \bottomrule
    \end{tabular}%
      \caption{The bound of each perturbation factor.}  \label{tab:ptb_parameter}%
       \vspace{-2mm}
\end{table}%

\subsection{Evaluating Semantic Perturbations}\label{sec:exp1}

In this experiment, we evaluate the effectiveness of the proposed objective function over different semantic perturbations and BEV models.
As summarised in Tab.~\ref{tab:ptb_parameter}, we introduce a parameter $\gamma$ to control the bound of perturbation factors. 
We constrain $\gamma$ to the range $[0.1, 0.4]$ for colour perturbations and use $\gamma \in [0.04, 0.1]$ for geometric perturbations, where $\gamma = 0.1$ allowing the perturbation to scale or translate the image by up to 10\% of its size.
The severity of motion blur is controlled by the size of the blur kernel~\cite{xie2023robobev}, and we set the kernel size to $\{5, 7, 9, 11\}$.  
Additionally, we bound the angle $\theta^{ang}\in [-\pi,\pi]$ and the direction $ \theta^{dir} \in [-1, 1]$ to explore the optimal kernel at different scales~\cite{eriba2019kornia}.

\begin{figure}[!tb]
    \centering
    \includegraphics[width=0.47\textwidth]{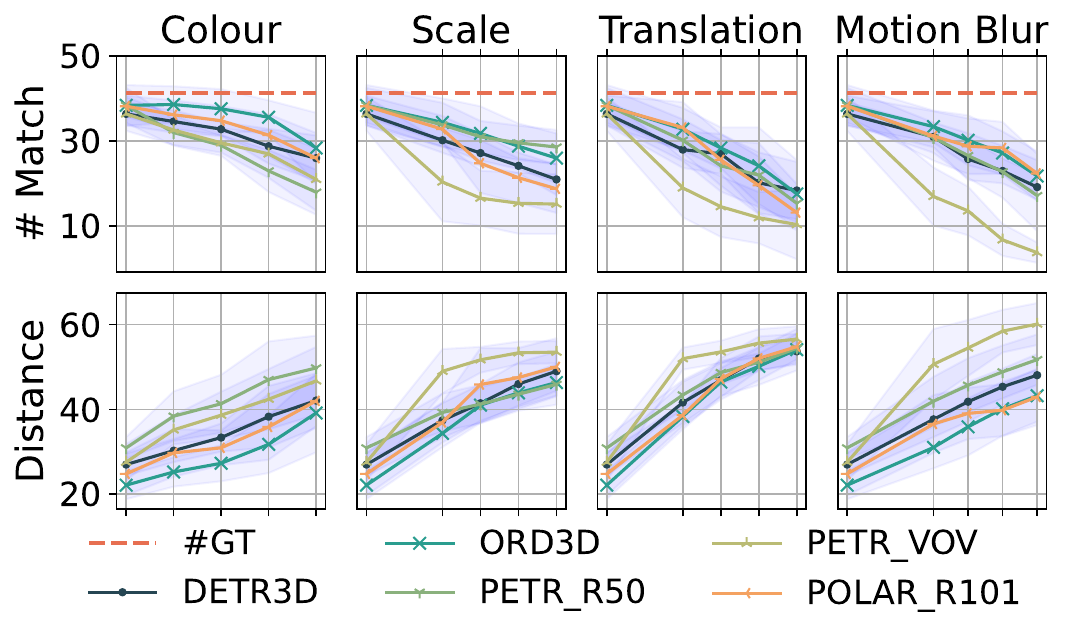}
    \caption{A illustration of the impact of semantic perturbations operated at different strengths on five BEV perception models. 
    The effects are quantified regarding the number of matches (\# Match) and the distance metric defined in Eq.~\eqref{eqn:final_object_func}. 
    Each figure starts with clean input frames, while we set the following perturbations to $\gamma \in [0.1, 0.4]$ for colour, $\gamma \in [0.04, 0.1]$ for geometry, and kernel sizes ${5, 7, 9, 11}$ for motion blur.
    The shaded area indicates the standard deviations of the models' performance.}
    \label{fig:exp_1}
     \vspace{-2mm}
\end{figure}

We uniformly sample 5 frames from the mini-validation set in nuScenes for evaluation and conduct 50 iterations of the SimpleDIRECT algorithm and record the obtained optimum results for the initial assessment of the impact of various perturbations. 
In Fig.~\ref{fig:exp_1}, we present a visualisation of the proposed objective function's value, namely the distance, and the corresponding number of matched bounding boxes under different perturbation setups.
While all types of perturbations could significantly reduce the target models' performance, the most notable observation is the strong negative correlation between the distance and the number of matches.
As the distance increases, the number of matched boxes decreases substantially. 
This phenomenon demonstrates that by maximising our proposed objective function, the generated adversarial semantic perturbations will effectively degrade the precision of the targeted BEV models.

\setlength{\tabcolsep}{0.4mm}
\begin{table}[tb]\small
  \centering
    \begin{tabular}{lccccccc}
    \toprule
    \multirow{2}[2]{*}{Methods} & \multicolumn{2}{c}{Colour} & \multicolumn{2}{c}{Scale \& Transl.} & \multicolumn{2}{c}{Motion Blur} &\multirow{2}[2]{*}{\makecell{Time \\(min.)}}\\
    \cmidrule(lr){2-3}\cmidrule(lr){4-5}\cmidrule(lr){6-7}
          & \#Match$\downarrow$ & Dist.$\uparrow$& \#Match$\downarrow$ & Dist.$\uparrow$& \#Match$\downarrow$ & Dist.$\uparrow$ \\
    \midrule
    \multicolumn{8}{c}{PETR RestNet50 } \\
    \midrule
    Random   & 24.4  & 37.1  & 19.6  & 41.6  & 23.6  & 39.4 & 8.2\\
    Natural+ & 28.8  & 40.3  & 19.2  & 52.6  & 36.0  & 34.3 & -\\
    Natural- & 23.4  & 43.4  & 21.4  & 51.4  & 36.2  & 34.9 & -\\
    \midrule
    DIRECT   & 24.0    & 45.9  & \textbf{13.8}  & 56.9  & \textit{22.4}  & 48.3 & 9.5 \\
    BO-EI     & \textbf{19.8}   & 48.5  & 15.6  & 54.1  & 23.2  & 46.4 & 55.1 \\
    \midrule
    $R=1$     & \textit{23.2}  & 46.9  & 14.6  & 55.4  & 23.4  & 46.6 & 9.0\\
    $R=2$     & \textit{23.2}  & 46.9  & 14.6  & 55.4  & \textbf{21.4}  & 48.5 & 9.2\\
    $R=3$     & \textit{23.2}  & 47.1  & \textit{14.4}  & 55.3  & \textbf{21.4}  & 48.5 & 9.6\\
    $R=4$     & \textit{23.2}  & 47.0  & \textit{14.4}  & 55.3  & \textit{22.4}  & 48.3 & 8.6\\
    \midrule
    \multicolumn{8}{c}{PETR VoVNet 800} \\
    \midrule
    Random   & 27.6  & 42.6  & 10.8  & 47.5  & 9.8  & 51.0 & 7.6\\
    Natural+ & 31.0    & 38.5  & 13.4  & 55.3  & 30.4  & 40.6 & -\\
    Natural- & 27.2  & 40.4  & 23.4  & 49.2  & 31.0  & 39.6 & -\\
    \midrule
    DIRECT   & 27.2  & 41.5  & 9.4   & 57.9  & \textit{7.0}  & 58.3 & 7.8\\
    BO-EI    & \textbf{24.8 }   & 44.4  & \textit{7.6}  & 58.7  & 9.6  & 56.6 & 29.0 \\
    \midrule
    $R=1$     & 27.4  & 39.7  & 10    & 57.2  & 9.2  & 57.4 & 7.6\\
    $R=2$     & 26.2  & 43.3  & \textbf{7.2}   & 59.0    & \textbf{6.8}  & 58.3 & 7.7\\
    $R=3$     & \textit{26.0}    & 43.7  & \textbf{7.2}   & 59.0    & \textit{7.0}  & 58.3  & 7.8\\
    $R=4$     & 26.8  & 42.4  & \textit{7.6}   & 58.8  & \textit{7.0}  & 58.3  & 7.6\\
    \bottomrule
    \end{tabular}%
  \caption{Comparison of SimpleDIRECT with baseline methods in terms of the number of matches (\# Matches), the distance metric in Eq.~\eqref{eqn:final_object_func}, and the average run time. 
  \textbf{SimpleDIRECT} is carried out with $R\in\{1,2,3,4\}$  to assess its performance.
  The best results are marked in \textbf{bold} and the second-best in \textit{italic}.}  \label{tab:exp2_comparsion}%
  \vspace{-2mm}
\end{table}%

\begin{table*}[htbp]\small
  \centering
    \begin{tabular}{cccccccccccccc}
    \toprule
    \multirow{2}[2]{*}{Perturbation} & \multirow{2}[2]{*}{Metric} & \multicolumn{4}{c}{BEVFormer} & \multicolumn{3}{c}{PETR} & \multicolumn{2}{c}{PolarFormer} & \multirow{2}[4]{*}{DETR3D} & \multirow{2}[4]{*}{ORA3D} & \multirow{2}[4]{*}{BEVDet} \\
    \cmidrule(lr){3-6}\cmidrule(lr){7-9}\cmidrule(lr){10-11}
          &       & small & small+tem. & base  & base+tem. & R50   & V800  & V1600 & R101  & VoV   &   &   &  \\
    \midrule
    \multirow{2}[2]{*}{None} & NDS$\uparrow$   & 0.279 & 0.399 & 0.357 & 0.426 & 0.321 & 0.363 & 0.402 & 0.408 & \textbf{0.471} & 0.375 & 0.410 & 0.356 \\
          & mAP$\uparrow$    & 0.189 & 0.355 & 0.319 & 0.294 & 0.298 & 0.341 & 0.380 & 0.367 & \textbf{0.429} & 0.313 & 0.359 & 0.300 \\
    \midrule
    \multirow{2}[2]{*}{Colour} & NDS$\uparrow$    & 0.223 & 0.291 & 0.261 & 0.223 & 0.162 & 0.234 & 0.263 & 0.323 & \textbf{0.409} & 0.306 & 0.350 & 0.000 \\
          & mAP$\uparrow$    & 0.108 & 0.231 & 0.187 & 0.187 & 0.095 & 0.155 & 0.184 & 0.282 & \textbf{0.329} & 0.227 & 0.276 & 0.000 \\
    \midrule
    \multirow{2}[2]{*}{Scale\&Shift} & NDS$\uparrow$    & 0.206 & 0.254 & 0.224 & 0.237 & 0.194 & 0.177 & 0.197 & 0.261 & \textbf{0.273} & 0.242 & 0.249 & 0.000 \\
          & mAP$\uparrow$    & 0.085 & \textbf{0.182} & 0.131 & 0.131 & 0.113 & 0.069 & 0.105 & 0.149 & 0.119 & 0.129 & 0.127 & 0.000 \\
    \midrule
    \multirow{2}[2]{*}{Motion Blur} & NDS$\uparrow$    & 0.179 & 0.231 & 0.246 & 0.229 & 0.170 & 0.081 & 0.189 & 0.326 & \textbf{0.367} & 0.268 & 0.283 & 0.000 \\
          & mAP$\uparrow$    & 0.100 & 0.159 & 0.180 & 0.160 & 0.104 & 0.010 & 0.093 & 0.229 & \textbf{0.272} & 0.169 & 0.199 & 0.000 \\
    \bottomrule
    \end{tabular}%
          \caption{Benchmarking BEV models against Semantic perturbations.}
  \label{tab:benchmark}%
\end{table*}%

\subsection{Ablation and Comparison}\label{sec:exp2}
The second part of our empirical study focuses on the effectiveness of the proposed SimpleDIRECT. 
The evaluation was done on the same sampled frames in Fig.~\ref{fig:exp_1}.
As targeted models, we select PETR~\cite{liu2022petr} with two backbones: ResNet50 and VoVNet 800. 
In terms of the number of matched bounding boxes, their average performance on the sampled frames is 38.4 and 36.4, respectively.
We perform three types of semantic perturbations, \ie, colour shift, motion blur, and the combination of scaling and translation.
Colour shift and motion blur are applied with $\gamma=0.3$ and a kernel size of 9, respectively, aligning with moderate severity levels in natural corruption~\cite{xie2023robobev}.
The combined geometric perturbation is conducted at $\gamma=0.1$ following~\citet{WangXRH23}'s practice.
Regarding baseline methods, as there are currently no other studies addressing the same scenario as ours, we adopt DIRECT~\cite{WangXRH23}, which is methodologically the most relevant work to ours, and Bayesian Optimisation with Expected Improvement acquisition function (\textbf{BO-EI}), the most popular black-box optimisation algorithm, for comparison. 
Additionally, inspired by the concept of natural corruption \cite{xie2023robobev}, we include the random search and evaluate each perturbation's performance at its maximum strength in both directions, using these as baselines denoted by \textbf{Natural$+/-$}.
All optimisation-based methods and the random search are allowed a maximum of 2000 queries for a fair comparison.

As shown in Tab.~\ref{tab:exp2_comparsion}, SimpleDIRECT was tested at $R \in \{1, 2, 3, 4\}$, with the best performance at $R = 3$.
There is a performance drop at $R = 1$ and $R = 4$, indicating that choosing too many or too few PO nodes affects efficiency.
Compared to baselines, optimisation-based approaches outperform random and natural corruptions, which highlights the limitation of relying on randomised and fixed perturbations for robustness evaluation, as such methods may not fully expose the vulnerabilities of the models.
SimpleDIRECT achieves the \textbf{best} or \textit{second-best} results across all perturbations, generally outperforming DIRECT when $R > 1$.
Although BO-EI, as a strong baseline, achieves the best performance on the colour shift perturbation, it was outperformed by DO methods elsewhere.
Regarding the runtime, DO methods take slightly longer than random search, where the runtime is mainly spent on querying the target model, while BO-EI is significantly slower than DO methods due to fitting the Gaussian process surrogate model in high-dimensional space~\cite{kawaguchi2015bayesian}.

\subsection{Benchmarking the Semantic Robustness}\label{sec:exp3}
In the previous sections, we develop a framework that can efficiently evaluate the adversarial robustness of BEV models against semantic perturbations, which enables us to benchmark the semantic robustness of recent camera-based BEV models.
Using the same perturbation setup as in the previous section, we conduct SimpleDIRECT at $R=3$ with a maximum of 2500 queries to find the optimal perturbation.
Following previous works~\cite{kong2023robo3d,xie2023robobev}, we evaluate robustness based on a model’s performance on perturbed inputs, with the highest-performing model considered the most robust.

Based on the mini-validation set of nuScenes, our benchmark includes ten BEV models, varying in scale, resolution, and backbones.
As summarised in Tab.~\ref{tab:benchmark},  the PolarFormer demonstrates the best robustness across different semantic perturbations.
The outstanding performance may be attributed to the polar coordinate system~\cite{jiang2023polarformer}, which potentially enhances the model's capability to comprehend geometric information from multi-view input.
Note that higher precision on clean frames does not always result in better robustness.
We can observe that ORA3D, PETR with VoVNet 1600, and PolarFormer with ResNet101 show similar performance on clean frames, but PolarFormer is notably more robust than the others.
In contrast, BEVDet demonstrated considerable resistance to natural corruptions~\cite{xie2023robobev}, but it fails to defend against semantic perturbations. 
This vulnerability may be attributed to the inherent data augmentation in BEVDet~\cite{huang2021bevdet}, which potentially weakens the model's reliability.
On the other hand, as the temporal information is not considered during the optimisation, we reuse the semantically perturbed images to evaluate the temporal version of BEVFormer. 
Both small and base versions show boosted performance on clean images with temporal information. 
However, the impact on robustness differs between versions: the small version benefits from temporal information and achieves improved robustness, while the robustness of the base version marginally decreases.

\subsection{A Case Study on the Full Validation Set}

Limited by the GPU resources, we could not afford to conduct the benchmark on the full validation set that contains 150 scenes (6019 frames).
In this case study, we extend our method to the full validation set by perturbing selected frames and subsequently applying the perturbation to other frames. 
Specifically, given that each scene contains approximately 40 frames, we perturb the initial frame of a scene and apply the resulting perturbation to the following nineteen frames. We then update the perturbation at the middle (21$^\text{th}$) frame and apply it to the remaining frames in the scene.
We adopt the small version of BEVFormer with and without using temporal information as two target models to study the impact of temporal information on the robustness performance.
Additionally, we conduct the random search with five attempts on all frames and report the best perturbation result on each frame as a baseline here to approximate the natural corruption in existing literature~\cite{zhu2023understanding,xie2023robobev}.
As evidenced in Tab.~\ref{tab:exp4_full_val}, despite the perturbations being optimised on only a small fraction (5\%) of the validation set, the proposed framework consistently outperforms randomised perturbation applied on the entire dataset.
While both models could resist random perturbations, the model with temporal information has a notably smaller performance drop under our optimised perturbation, compared to when the temporal information is disabled.

\setlength{\tabcolsep}{0.8mm}
\begin{table}[!tb]\small
  \centering
    \begin{threeparttable}
    \begin{tabular}{cccccccc}
    \toprule
    \multirow{2}[2]{*}{Temp.} & \multirow{2}[2]{*}{Methods} & \multicolumn{2}{c}{Colour} & \multicolumn{2}{c}{Scale \& Transl.} & \multicolumn{2}{c}{Motion Blur} \\
    \cmidrule(lr){3-4}\cmidrule(lr){5-6}\cmidrule(lr){7-8}
          &       & NDS$\downarrow$   & mAP$\downarrow$   & NDS$\downarrow$   & mAP$\downarrow$   & NDS$\downarrow$   & mAP$\downarrow$ \\
    \midrule 
    \multirow{2}[4]{*}{\textbf{W/O}\tnote{1}} & Rnd. & 0.252 & 0.122 & 0.244 & 0.112 & 0.247 & 0.120 \\
\cmidrule{2-8}  & Ours  & \textbf{0.101} & \textbf{0.007} & \textbf{0.074} & \textbf{0.006} & \textbf{0.103} & \textbf{0.007} \\
    \midrule 
    \multirow{2}[4]{*}{\textbf{W/}\tnote{2}} & Rnd. & 0.373 & 0.307 & 0.338  & 0.258 & 0.370  & 0.299 \\
\cmidrule{2-8}   & Ours  & \textbf{0.347} & \textbf{0.267} & \textbf{0.286} & \textbf{0.181} & \textbf{0.340} & \textbf{0.255} \\
    \bottomrule
    \end{tabular}%
    \begin{tablenotes}
    \item[1] Performance on clean frames: NDS=0.263, mAP=0.132
    \item[2] Performance on clean frames: NDS=0.479, mAP=0.370
    \end{tablenotes}
      \end{threeparttable}
  \caption{A case study on extending our framework to the full validation set. We perturb the 1$^\text{st}$ and 21$^\text{th}$ frames in each scene (300 frames in total) and subsequently apply the perturbation to the following frames. 
  }
  \label{tab:exp4_full_val}%
  \vspace{-2mm}
\end{table}%

\section{Conclusion}

In this work, we propose a query-based black-box framework for evaluating the `worst-case' robustness of BEV detection models against semantic perturbations.
In our experiment, this method outperforms natural perturbation and strong optimisation-based baselines and significantly reduces the precision of camera-based BEV models, demonstrating the effectiveness of the proposed framework for measuring the vulnerability of existing BEV models.
As autonomous driving technology continues to evolve, 
the proposed evaluation framework can be easily deployed on newly developed perception models due to its black-box nature.
However, our robustness evaluation is still limited to certain semantic perturbations.
Developing more practical and representative perturbations offers a promising avenue for future research.

\section{Acknowledgments}
FW is funded by the Faculty of Environment, Science and Economy at the University of Exeter. WR and XH are the corresponding authors.
The work was undertaken during FW’s internship at the TACPS Lab at the University of Liverpool, supported by the U.K.
EPSRC through End-to-End Conceptual Guarding of Neural
Architectures [EP/T026995/1]. XH is also partially supported by the same project. 
We would like to thank the anonymous reviewers for their insightful and constructive comments, and we are also grateful to Jinwei Hu for his assistance at the early stage of this project.

\bibliography{aaai25}
\clearpage

\section{Appendix}

\subsection{More Details about SimpleDIRECT}\label{app:opt_algo}
In this section, we explain more about the space trisection process in DIRECT and discuss the convergence of our proposed improvement, SimpleDIRECT.

\subsubsection{Visualising the Space Division Process}\label{app:visual_division}
\begin{figure}[ht]
    \centering
    \includegraphics[width=0.9\linewidth]{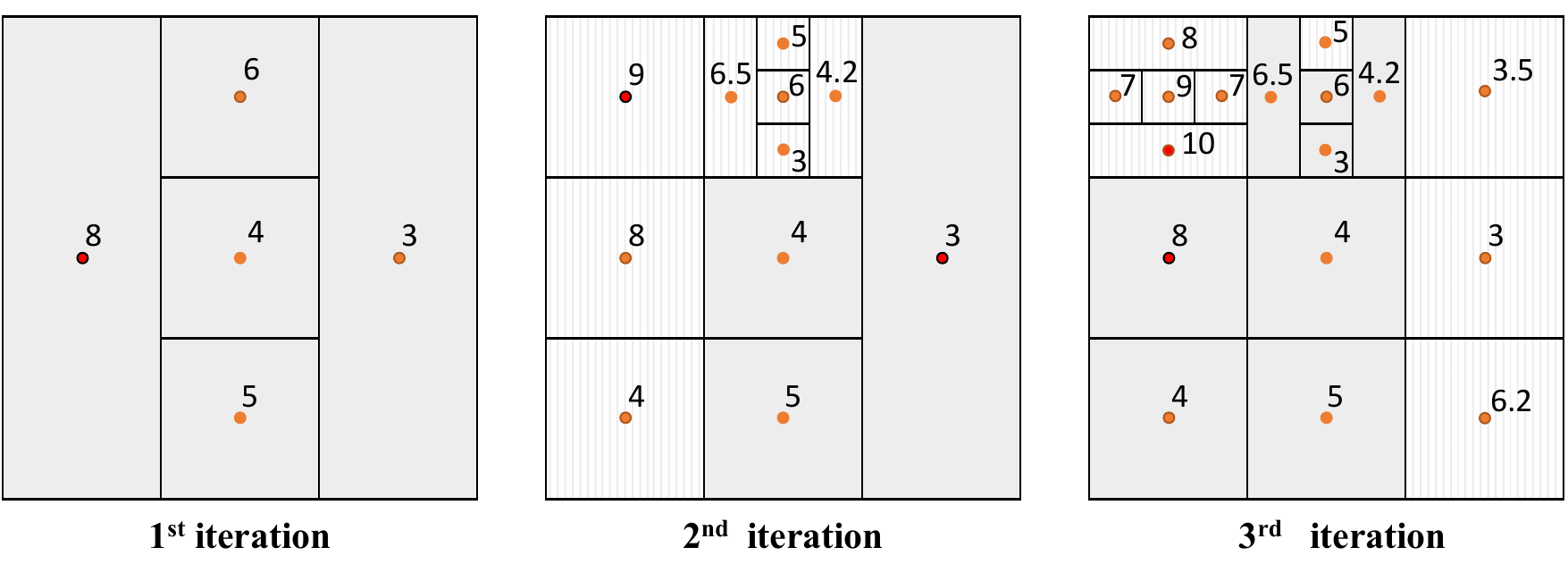}
     \caption{An illustration of the first three iterations of solving a 2-D maximisation problem. }
    \label{fig:division}
\end{figure}

In Fig.~\ref{fig:division}, we illustrate the space trisection process across the first three iterations. 
The algorithm identifies potential optimal subspaces for trisection during each iteration, methodically exploring the selected areas.
While the original DIRECT was proposed to solve minimisation problems, we adopted it to handle a maximisation problem, where the slopes computed in Fig.~\ref{fig:simple_cond} are more likely to be negative.

\subsubsection{Convergence}\label{app:proof}
Considering a continuous objective function, deterministic optimisation is guaranteed to converge to the global optimum within a finite number of queries if the following two conditions are met~\cite{Gablonsky01}: \textit{i)} there is at least one node that will be identified as the potential optimal node; \textit{ii)} Every node will be subdivided after finite iterations.
As elaborated in Lemma~\ref{lem:find_po}, a node $\Theta_p$ subjecting to $\mathcal{I}^h_2 = \varnothing$ and Eq.~\eqref{eqn:po_cond0} always satisfies Eq.~\eqref{eqn:po_cond1} and~\eqref{eqn:po_cond2}.
This is because the right-hand side of Eq.~\eqref{eqn:po_cond1},
$$
\max _{\Theta_q \in \mathcal{I}^h_{2}} \frac{\mathcal{L}\left(\theta_{q}\right)-\mathcal{L}\left(\theta_{p}\right)}{\delta_{q}-\delta_{p}},
$$
could potentially reach $-\infty$ in the absence of reference nodes in $\mathcal{I}^h_2$.
Therefore, at each iteration, DIRECT can select at least one node with the largest diameter as the potential optimal node, ensuring its convergence.
On the other hand, the proposed SimpleDIRECT selects the node with the largest diameter subsequent to the refinement process defined by Eq.\eqref{eqn:po_cond0} and Eq.\eqref{eqn:po_cond2}.
Such a node selection strategy preserves the convergence conditions consistent with those of DIRECT, resulting in the query complexity bound described in Theorem~\ref{thm:precison_bound}.

\begin{theorem}[\citet{WangXRH23}]\label{thm:precison_bound}
Slightly abusing the notation, let $\Theta$ be the $n$-dimensional united search space and $\tilde{K}$ be the Lipschitz constant of the objective function $L$ within $\Theta$.
The gap between current maxima and global maxima after $T$ iterations is bounded, which can be written as
\begin{equation}\label{eqn:thm_1}
    ||\mathcal{L}_{\max} - \max _{\theta \in \Theta} \mathcal{L}(\theta) ||_1 \leq \varepsilon < \tilde{K}\cdot(T+1)^{-\frac{1}{n}}.
\end{equation}
Therefore, to achieve any desired $\varepsilon$, the algorithm needs to be carried out up to $\mathcal{O}\big((\tilde{K}/\varepsilon)^n\big)$ iterations.
\end{theorem}

\subsection{Implementation Details}
In this section, we introduce the implementation details of this work and present additional ablation studies to justify the selection of hyperparameters.
The BEV models are adopted from RoboBEV~\cite{xie2023robobev}, and our code will be made publicly accessible upon acceptance of this work.

\subsubsection{Hardware}
All experiments are performed on a desktop with an Intel i7-10700KF CPU, an RTX 3090 GPU, and 48 gigabytes of memory.

\subsubsection{Numerical Examples}
In Fig.~\ref{fig:po_partition}, we perform DIRECT on a 2D Ackley function for 4 iterations and visualise the landscape of the function value.
In Fig.~\ref{fig:po_cond}, we conduct DIRECT with $\epsilon=0.01$ on a 6D Schwefel function and record the number of nodes qualified by different conditions.

\subsubsection{Perturbation}
In the experimental section, we use Kornia~\cite{eriba2019kornia} to perform colour shifting and motion blur perturbations, and the Spatial Transformer Network (STN)~\cite{JaderbergSZK15} is adopted to conduct scaling and translation perturbations.
Additionally, since each frame in the nuScenes dataset comprises six images, optimise colour shift, scaling and translation, and motion blur perturbations raise 18-D, 24-D, and 12-D adversary spaces, respectively.

\begin{figure*}[htb]
    \centering
    \includegraphics[width=0.8\textwidth]{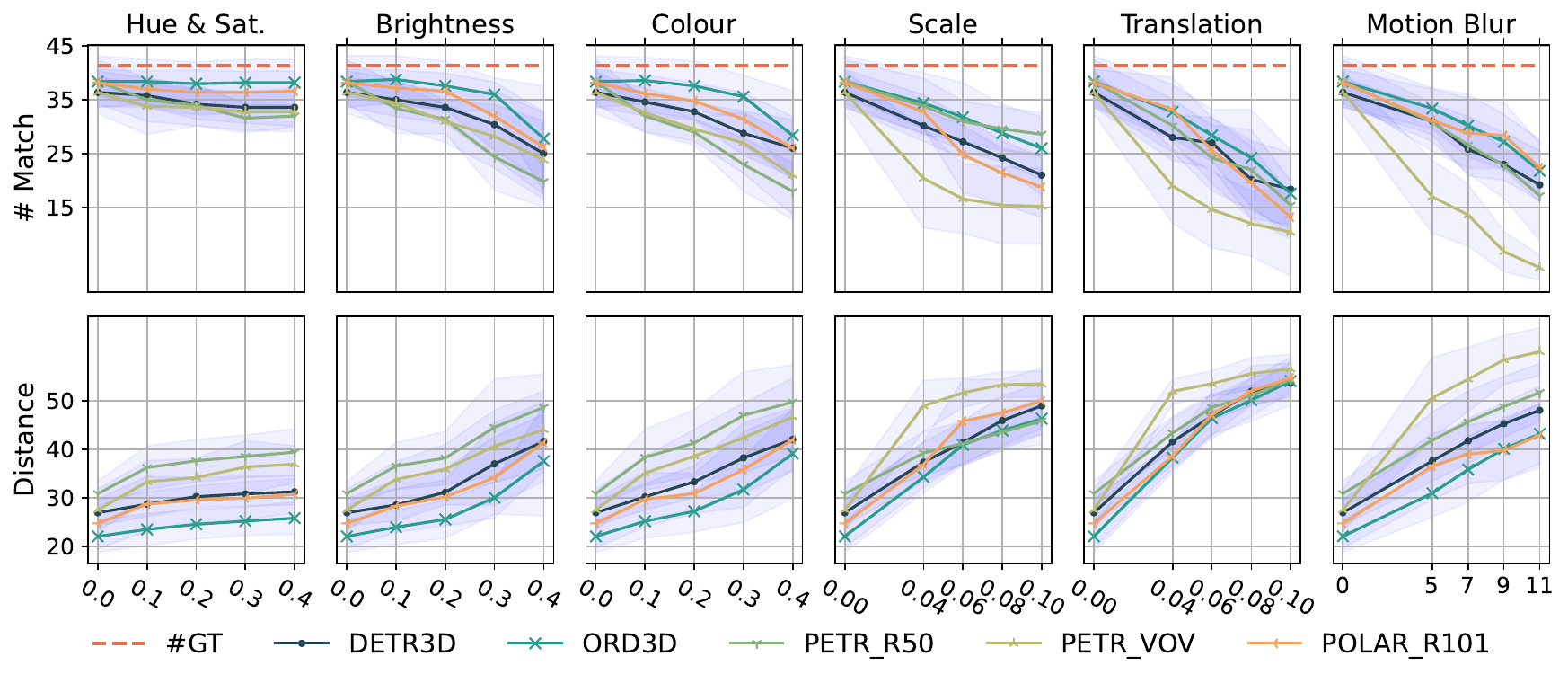}
    \caption{A illustration of the impact of semantic perturbations operated at different strengths on five BEV perception models. The effects are quantified in terms of the number of matches (\# Match) and the distance metric defined in Eq.~\eqref{eqn:final_object_func}. 
    The dashed red line represents the average number of ground truth boxes (\# GT) across the sampled frames.
    The standard deviation of the models' performance is indicated by the shaded area surrounding each line.}
    \label{fig:exp_1_detailed}
\end{figure*}

Same as in Fig.~\ref{fig:exp_1}, where colour and geometric perturbations are carried out with $\gamma \in [0.1, 0.4]$ and $\gamma \in [0.04, 0.1]$, respectively and the kernel size of motion blur is fixed at $\{5, 7, 9, 11\}$, We provide a more detailed illustration of the impact of each semantic perturbation.
As shown in Fig.~\ref{fig:exp_1_detailed}, although we conduct the colour shift as a combination of perturbations on hue, saturation, and brightness, brightness appears to be the most influential factor.
This phenomenon could indicate that the tested BEV models are sensitive to the lighting conditions.

\begin{figure*}[!ht]
     \centering
     \begin{subfigure}[c]{0.45\textwidth}
         \centering
         \includegraphics[width=\textwidth]{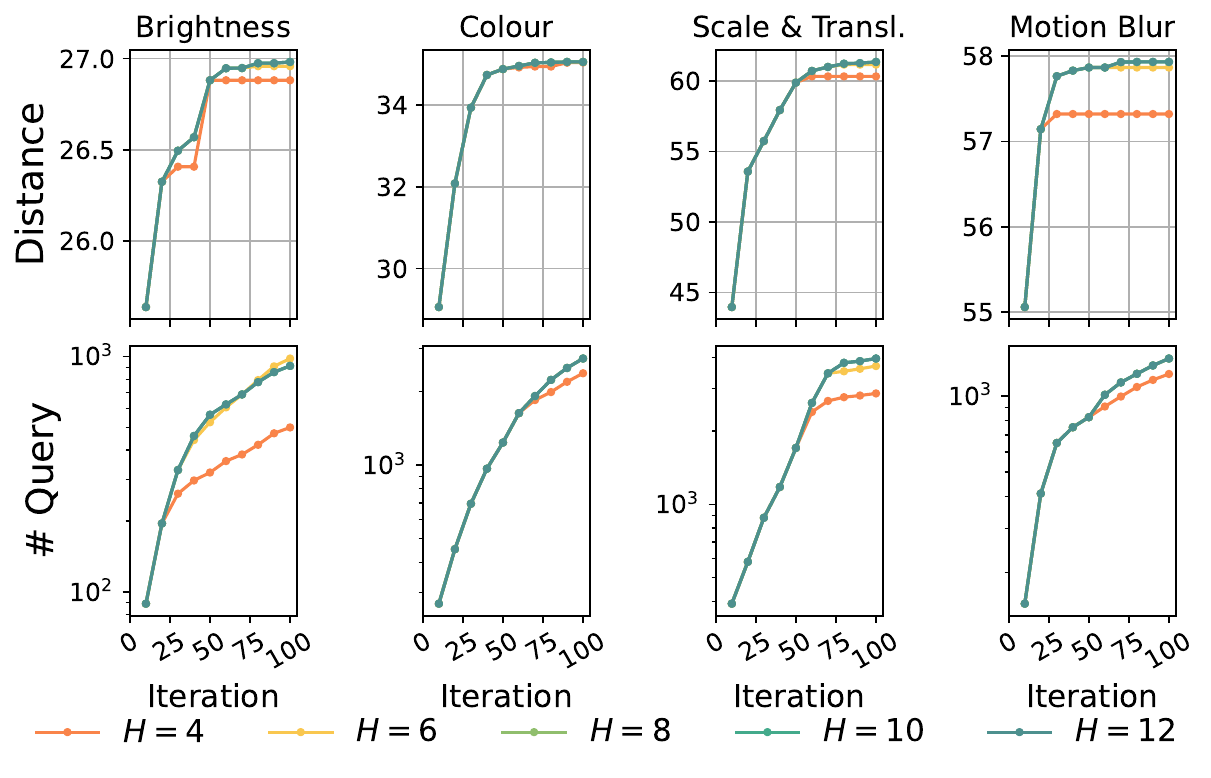}
         \caption{ResNet50}
         \label{fig:exp2_r50}
     \end{subfigure}
     \hfill
     \begin{subfigure}[c]{0.45\textwidth}
         \centering
         \includegraphics[width=\textwidth]{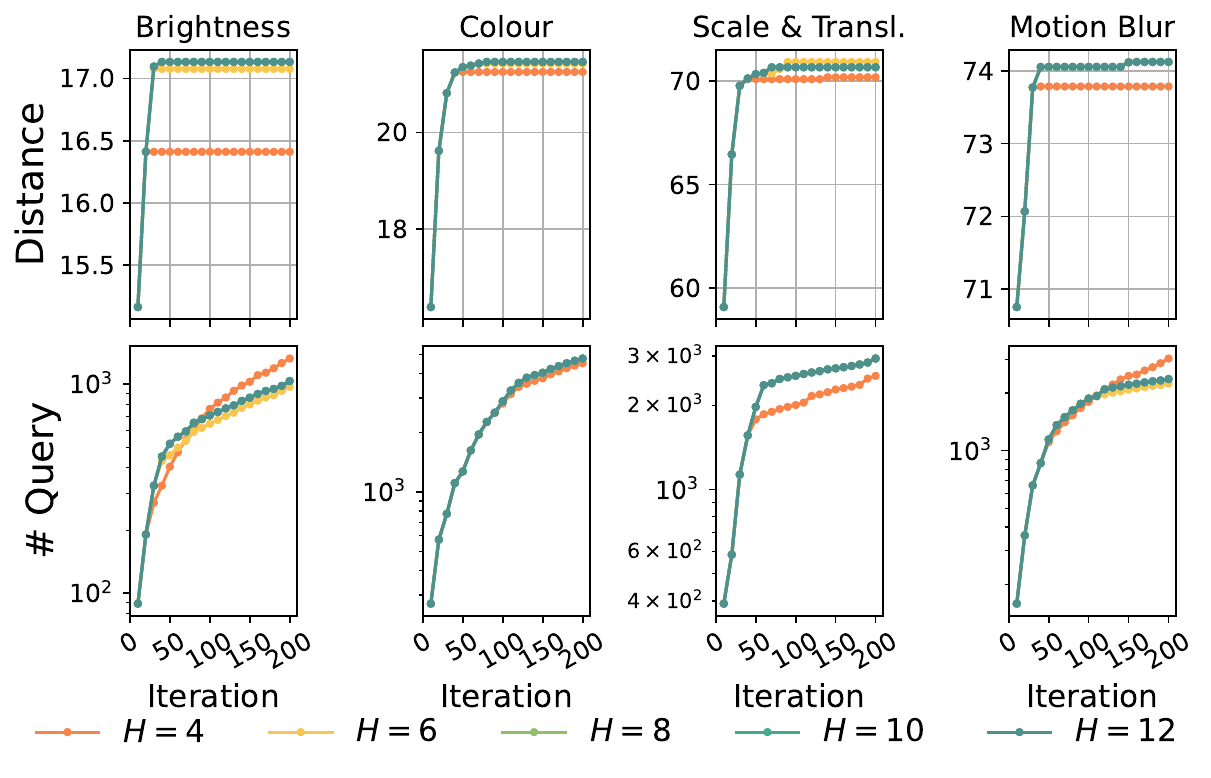}
         \caption{VoVnet $800\times 320$}
         \label{fig:exp2_vov}
     \end{subfigure}
        \caption{A showcase of conducting DIRECT at depth $H \in \{4,6,8,10,12\}$ on four semantic perturbations.
        The intensity of brightness and colour perturbations is carried out at $\gamma = 0.3$, while both geometric perturbations are upper bounded by $\gamma = 0.1$, and the kernel size of motion blur perturbation is fixed at 9.}
        \label{fig:exp_2}
          \vspace{-3mm}
\end{figure*}

\subsubsection{Determining the Depth}
To find the most suitable depth $H$ for our evaluation, we run DIRECT at depth $H\in \{4,6,8,10,12\}$ on PETR with ResNet50 and VoVnet backbones, respectively.
It can be seen from Fig.~\ref{fig:exp_2} that the optimisation performance at $H=4$ is generally worse than other settings, while $H=6$ seems to be an ideal option that requires fewer queries and offers comparable results in terms of maximising the distance metric.
Therefore, based on our observation from Fig.~\ref{fig:exp_2}, both DIRECT and SimpleDIRECT are carried out at depth $H=6$ in the comparison presented in Tab.~\ref{tab:exp2_comparsion} and the robustness benchmark summarised in Tab.~\ref{tab:benchmark}.

\begin{table*}[htb]\small
  \centering
    \begin{tabular}{cccccccccc}
    \toprule
    \multirow{2}[2]{*}{Backbone} & \multirow{2}[2]{*}{Obj. Func.} & \multicolumn{2}{c}{Brightness} & \multicolumn{2}{c}{Colour } & \multicolumn{2}{c}{Scale \& Transl. } & \multicolumn{2}{c}{Motion Blur } \\
    \cmidrule(lr){3-4}\cmidrule(lr){5-6}\cmidrule(lr){7-8}\cmidrule(lr){9-10}
          &       & \# Match$\downarrow$& Distance$\uparrow$& \# Match$\downarrow$ & Distance$\uparrow$& \# Match$\downarrow$ & Distance$\uparrow$& \# Match$\downarrow$ & Distance$\uparrow$\\
    \midrule\midrule
    \multicolumn{1}{l}{\multirow{2}[2]{*}{ResNet}} 
    &  Eq.~\eqref{eqn:final_object_func}     &   \textbf{23.6}   &  45.7     & 23.2  & 47.1  & \textbf{14.4}      &     55.3  & \textbf{21.4}  & 48.5 \\
          & Eq.~\eqref{eqn:det_object_func}     &    24    &  45.1     &  \textbf{23}     & 46.7       &   17.2    &     55.3  &   \textbf{21.4}    &  48.1\\
    \midrule\midrule
    \multicolumn{1}{l}{\multirow{2}[2]{*}{VoVnet}}  
    & Eq.~\eqref{eqn:final_object_func}     & \textbf{28.2}  & 40.5  & \textbf{26}    & 43.7  & \textbf{7.2}   & 59   & \textbf{7.0}  & 58.3 \\
          & Eq.~\eqref{eqn:det_object_func}     & 28.4  & 39.4  & 27 & 41.5  & 7.6   & 57.9  & 7.8    & 58.3\\
    \bottomrule
    \end{tabular}%
      \caption{Comparing the objective functions given in Eq.~\eqref{eqn:final_object_func} and~\eqref{eqn:det_object_func}.
  Same as in Tab.~\ref{tab:exp2_comparsion}, the targeted models are PETR with ResNet and VoVnet backbones.
  The optimisation algorithm here is SimpleDIRECT with $R=3$, and the computational budget is up to 2500 queries.}  \label{tab:exp2_det}%
  \vspace{-3mm}
\end{table*}%

\subsection{Discussion about the Objective Function}
In the context of autonomous driving, the widely used performance metric, NDS~\cite{caesar2020nuscenes}, evaluates BEV models based on the 2D centre distances between predicted and ground truth bounding boxes rather than the IoU metric. 
In line with this, we propose a distance-based objective function given by 
\begin{equation*}\textstyle
  \mathcal{L}\big(F(S_\theta(\bfx)), \bfy\big) =  \sum_{v=1}^V \min\big(\min \mathcal{D}(\hat{\bfy}^v_{\theta}, \bfy_v), \tau\big),\tag{3}
\end{equation*}
where $C(\cdot)$ computes the 2D centre distances between input bounding boxes and $\tau$ is a distance threshold.
Maximising the proposed objective function aims to increase the distance between matched bounding boxes while also preventing the creation of new matches.
As demonstrated in Fig.~\ref{fig:exp_1}, the perturbations optimised based on our objective function show a strong negative impact on BEV models, satisfying the need to conduct adversarial robustness evaluation.

While other factors like classification score and box size could potentially be integrated into the objective function, we argue that their importance is secondary.
This is because their impact becomes negligible if the detected boxes are dismissed for falling outside the distance threshold.
For example, considering the following objective function
\begin{equation}\label{eqn:det_object_func}
\begin{aligned}
    &\mathcal{L}\big(\theta; F, S, \bfy) \\
    &\quad= \sum_{v=1}^V \big(\min\big(\min C(\hat{\bfy}^v_{\theta}, \bfy_v), \tau\big) - C_{cls}(\hat{\bfy}^v_{\theta}, \bfy_v)\big),
\end{aligned}
\end{equation}
where $C_{cls}()$ denotes a function that returns the classification score of the matched boxes. 
We implement the objective function given in Eq.~\eqref{eqn:det_object_func}, which could maximise the distance and diminish the classification score simultaneously, and make a comparison to the proposed distance-based objective given in Eq.~\eqref{eqn:final_object_func}.
As shown in Tab.~\ref{tab:exp2_det}, we can see that minimising the classification score of matched boxes does not enhance attack performance in 7 out of 8 settings. 
This suggests that optimising classification scores has a limited effect on improving attacks.
While we do believe that more criteria could be included in the proposed framework, this topic seems to be out of the scope of a single paper and could be investigated in future works.

\subsection{Additional Experiments}
\subsubsection{More Comparison with DIRECT}
The proposed SimpleDIRECT is a new deterministic optimisation algorithm that could be used in a wide range of tasks.
To evaluate the generalisability of SimpleDIRECT, we compared its performance to DIRECT under the same setting in~\citet{WangXRH23}, which evaluated the robustness of ImageNet classifiers against geometric transformation attacks.
The comparison is made on five models when subjects to a combined perturbation, including rotation, scaling, and translation.
It can be seen from Tab.~\ref{tab:add_experiment} that SimpleDIRECT outperforms DIRECT in terms of the Attack Success Rate (ASR) and run time.
Please note that these image classifiers allow for parallel queries, making them more efficient than BEV models, which can only query sequentially~\cite{xie2023robobev}. 
Due to the significantly reduced time cost when querying the target models, the improvement in runtime is more observable here than in Tab.~\ref{tab:exp2_comparsion}.
These results provide further empirical evidence that the improvements made in SimpleDIRECT are substantial and generalisable.

\setlength{\tabcolsep}{1.5mm}
\begin{table}[!ht]\small
  \centering
    \begin{tabular}{lcccc}
    \toprule
    \multicolumn{1}{c}{\multirow{2}[4]{*}{Model}} & \multicolumn{2}{c}{DIRECT} & \multicolumn{2}{c}{SimpleDIRECT} \\
\cmidrule{2-5}          & ASR & Time (s) & ASR & Time (s) \\
    \midrule
    ResNet50 & 39.29\% & 5.0$\pm$0.5 & \textbf{39.54\%} & \textbf{3.9$\pm$0.4} \\
    W.ResNet50 & \textbf{38.24\%} & 6.0$\pm$0.6 & \textbf{38.24\%} & \textbf{5.0$\pm$0.5} \\
    Vit$_{16\times16}$  & 47.91\% & 4.9$\pm$0.8 & \textbf{48.16\%} & \textbf{3.8$\pm$0.7} \\
    Large Beit$_{16\times16}$ & 22.90\% & 9.2$\pm$1.2 & \textbf{23.13\%} & \textbf{8.3$\pm$1.2} \\
    Swin  & 55.11\% & 5.8$\pm$0.5 & 55.11\% & \textbf{4.7$\pm$0.4} \\
    \bottomrule
    \end{tabular}%
      \caption{Comparison between DIRECT and SimpleDIRECT}  \label{tab:add_experiment}%
        \vspace{-3mm}
\end{table}%

\subsubsection{Case Study on the Full Validation Set}
The PolarFormer with VoVNet backbone demonstrated the best robustness in our benchmark.
Therefore, we evaluate its performance on the full validation set following the same perturbation strategy in Tab.~\ref{tab:exp4_full_val}.
As shown in Tab.~\ref{tab:exp4_full_val_polar}, while the randomised perturbations only marginally degrade the model’s performance, our evaluation method significantly reduces its mAP score, leading to a notable decrease in its NDS scores.
Combining with the results in Tab.~\ref{tab:exp4_full_val}, it appears that both BEVFormer-small and PolarFormer with VoVNet only show marginal performance drop when subject to randomised semantic perturbation but are considerably vulnerable to adversarially optimised perturbation.
In contrast, BEVFormer-small with temporal information is relatively more sensitive to randomised semantic perturbation but shows better robustness to optimised perturbation.
This phenomenon suggests that the robustness of BEV models could potentially benefit from incorporating temporal information, though further investigation is required for confirmation. 
This appears to be an interesting direction for future research. 
Given that the primary contribution of this work is the proposal of a black-box evaluation framework, we believe that exploring this direction falls outside the scope of this paper and will be addressed in our future works.

\setlength{\tabcolsep}{0.8mm}
\begin{table}[!tb]\small
  \centering
    \begin{threeparttable}
    \begin{tabular}{cccccccc}
    \toprule
    \multirow{2}[2]{*}{Model} & \multirow{2}[2]{*}{Methods} & \multicolumn{2}{c}{Colour} & \multicolumn{2}{c}{Scale \& Transl.} & \multicolumn{2}{c}{Motion Blur} \\
    \cmidrule(lr){3-4}\cmidrule(lr){5-6}\cmidrule(lr){7-8}
          &       & NDS$\downarrow$   & mAP$\downarrow$   & NDS$\downarrow$   & mAP$\downarrow$   & NDS$\downarrow$   & mAP$\downarrow$ \\
    \midrule 
    \multirow{2}[2]{*}{\makecell{Polar\tnote{1}\\VoV}} & Random & 0.531 & 0.460 & 0.517 & 0.429 & 0.542 & 0.472 \\
\cmidrule{2-8}  & Ours  & \textbf{0.253} & \textbf{0.019} & \textbf{0.164} & \textbf{0.014} & \textbf{0.280} & \textbf{0.021} \\
    \bottomrule
    \end{tabular}%
    \begin{tablenotes}
    \item[1] Performance on clean frames: NDS=0.562, mAP=0.500
    \end{tablenotes}
      \end{threeparttable}
  \caption{We perturb the 1$^\text{st}$ and 21$^\text{th}$ frames in each scene (300 frames in total) and subsequently apply the perturbation to the following frames. 
  }
  \label{tab:exp4_full_val_polar}%
    \vspace{-3mm}
\end{table}%

\subsection{Perturbation Visualisation}
\label{app:visal}
\begin{figure*}[!ht]
     \centering
    \includegraphics[width=0.9\linewidth]{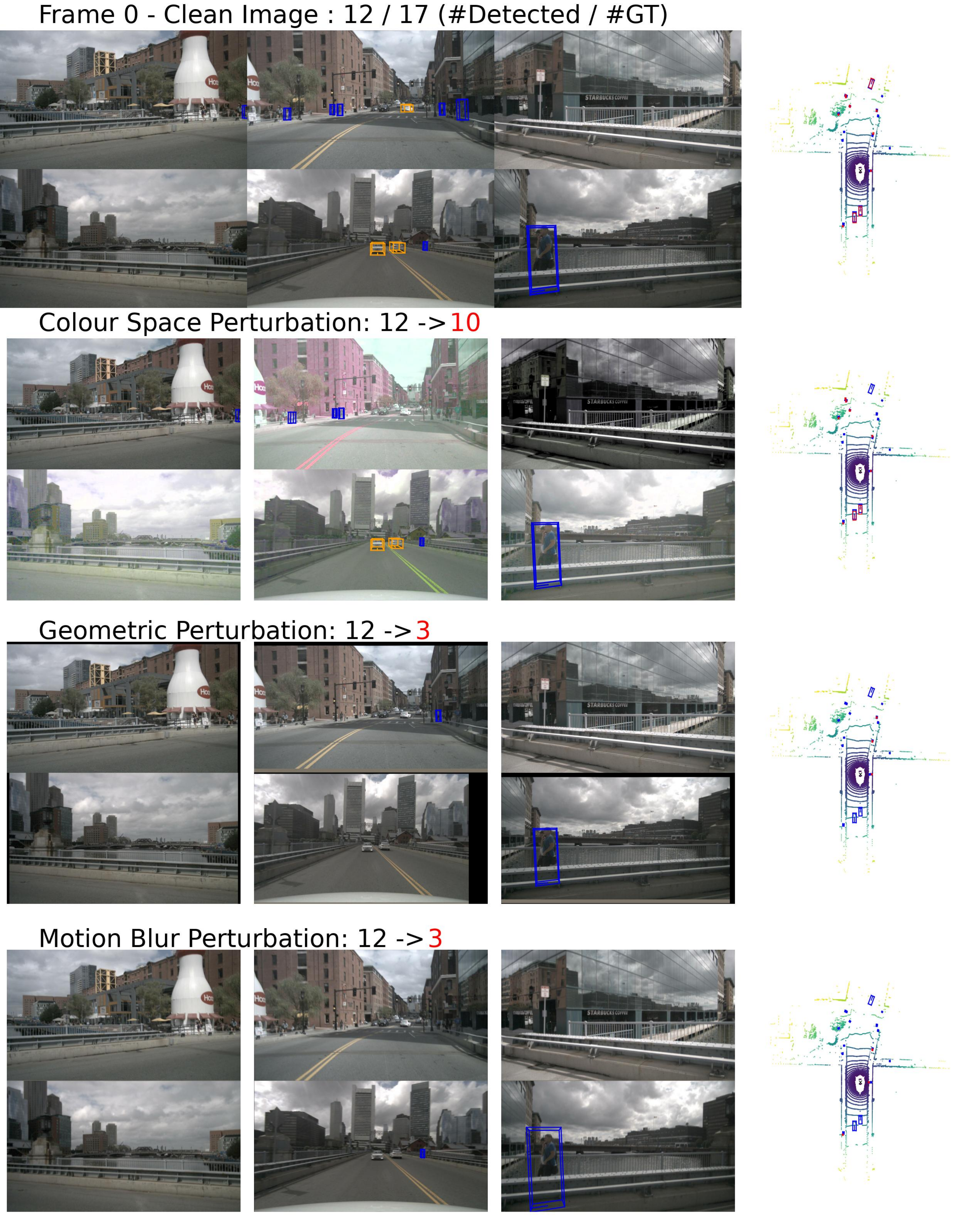}
        \caption{A illustration of the perturbed frames and the BEV perceptions produced by DETR3D.}
        \label{fig:all_bev}
\end{figure*}

In this section, we provide a set of visualisations to show the effect of each perturbation considered in Tab.~\ref{tab:benchmark}.
We reproduce the perturbations on the first frame in the validation set and adopt DETR3D to generate BEV preception.
The first row in Fig.~\ref{fig:all_bev} shows the model's performance on the clean images, while the following three rows visualise the colour, the combination of scaling and translation, and motion blur perturbations.
In the BEV predictions on the right-hand side, we use blue boxes to represent the ground truth bounding boxes, and the red boxes are the model's predictions.
Colour shift perturbation is the most noticeable one, but it does not obscure the semantic content for human observers. 
In contrast, while perceptible, scaling and translation perturbations substantially diminish the model's accuracy, even though objects still appear within the images. 
The motion blur perturbation, which is nearly imperceptible to humans, successfully deceives the targeted DETR3D model as well.
To further illustrate the proposed evaluation, we have included one video demonstration of each perturbation in the supplementary material.

\end{document}